\documentclass[runningheads]{llncs}
\usepackage{graphicx}
\usepackage{comment}
\usepackage{amsmath,amssymb} 
\usepackage{color}

\usepackage{multirow}
\usepackage{makecell}
\usepackage{subcaption}
\usepackage[font=small,skip=4pt]{caption}
\usepackage{pifont}
\usepackage{eso-pic}
\usepackage{xspace}

\usepackage[pagebackref=true,breaklinks=true,letterpaper=true,colorlinks,bookmarks=false]{hyperref}

\newcommand{\mcth}[2]{\multicolumn{#1}{c|}{\thead{#2}}}
\newcommand{\mrc}[2]{\multirowcell{#1}{#2}}

\newcommand{\cmark}{\ding{51}}
\newcommand{\xmark}{\ding{55}}

\newcommand{\mobilenet}[1]{MobileNet-v#1}
\newcommand{\mobilenetsuffix}[1]{-v#1}

\graphicspath{{./figures/}}
\makeatletter
\DeclareRobustCommand\onedot{\futurelet\@let@token\@onedot}
\def\@onedot{\ifx\@let@token.\else.\null\fi\xspace}
\def\eg{e.g\onedot}

\def\etal{\emph{et al}\onedot}
\makeatother

\begin{document}
\pagestyle{headings}
\mainmatter

\title{MeliusNet: Can Binary Neural Networks Achieve MobileNet-level Accuracy?} 

\titlerunning{MeliusNet}
\author{Joseph Bethge\inst{1} \and
Christian Bartz\inst{1} \and
Haojin Yang\inst{1,2} \and
Ying Chen\inst{2} \and
Christoph Meinel\inst{1}}
\authorrunning{J. Bethge et al.}
\institute{Hasso Plattner Institute, University of Potsdam, Germany\\
\email{\{joseph.bethge,christian.bartz,haojin.yang,christoph.meinel\}@hpi.de} \and
AI Labs, Alibaba Group \email{\{haojin.yhj,chenying.ailab\}@alibaba-inc.com}}
\maketitle

\begin{abstract}

\noindent
Binary Neural Networks (BNNs) are neural networks which use binary weights and activations instead of the typical 32-bit floating point values.
They have reduced model sizes and allow for efficient inference on mobile or embedded devices with limited power and computational resources.
However, the binarization of weights and activations leads to feature maps of lower quality and lower capacity and thus a drop in accuracy compared to traditional networks.
Previous work has increased the number of channels or used multiple binary bases to alleviate these problems.
In this paper, we instead present an architectural approach: MeliusNet.
It consists of alternating a DenseBlock, which increases the feature capacity, and our proposed ImprovementBlock, which increases the feature quality.
Experiments on the ImageNet dataset demonstrate the superior performance of our MeliusNet over a variety of popular binary architectures with regards to both computation savings and accuracy.
Furthermore, with our method we trained BNN models, which for the first time can match the accuracy of the popular compact network MobileNet-v1 in terms of model size, number of operations and accuracy.
Our code is published online:
\\ \url{https://github.com/hpi-xnor/BMXNet-v2}
 \end{abstract}

\section{Introduction}
\label{sec:introduction}

\noindent
The success of deep convolutional neural networks in a variety of machine learning tasks, such as image classification \cite{He2017,krizhevsky2012imagenet}, object detection \cite{YOLOv1,Faster-rcnn}, text recognition \cite{jaderberg_2014_eccv}, and image generation \cite{wassersteingan,goodfellow2014generative}, has led to the design of deeper, larger, and more sophisticated neural networks.
However, the large size and high number of operations of these accurate models severely limit the applicability on resource-constrained platforms, such as those associated with mobile or embedded devices.
There are many existing works aiming to solve this problem by reducing memory requirements and accelerating inference.
These approaches can be roughly divided into a few research directions:
knowledge distillation \cite{crowley2018moonshine,polino2018model,tung2019similarity},
network pruning techniques \cite{Han2015,han2015learning},
compact network designs \cite{Howard_2019_ICCV,Howard2017,Iandola2016,Sandler_2018_CVPR,Zhang2017},
and low-bit quantization \cite{courbariaux2015binaryconnect,Rastegari2016,Zhou2016}, wherein the full-precision 32-bit floating point weights (and in some cases also the activations) are replaced with lower-bit representations, e.g. 8 bits or 4 bits.
The extreme case, Binary Neural Networks (BNNs), was introduced by \cite{Courbariaux2016,Rastegari2016} and uses only 1 bit for weights and activations.

It was shown in previous work that the BNN approach is especially promising, since a binary convolution can be sped up by a factor higher than 50$\times$ while using only less than 1\% of the energy compared to a 32-bit convolution on FPGAs and ASICs \cite{Mishra2017}.
This speed-up can be achieved by replacing the multiplications (and additions) in matrix multiplications with bit-wise \texttt{xnor} and \texttt{bitcount} operations \cite{Mishra2017,Rastegari2016}, processing up to 64 values in one operation.
However, BNNs still suffer from accuracy degradation compared to their full-precision counterparts \cite{Gu_2019_ICCV,Rastegari2016}.
To alleviate this issue, there has been work to approximate full-precision accuracy by using multiple weight bases \cite{Lin2017,Zhuang_2019_CVPR} or increasing the channel number in feature maps \cite{Mishra2017,Shen2019a}.
We briefly review the related work in more detail in \autoref{sec:related-work}.

Furthermore, alternative approaches to BNNs, such as the compact network structure \mobilenet{1} \cite{Howard2017} have achieved higher accuracy in the past.
More recent work on compact network structures, such as \mobilenet{2} or \mobilenetsuffix{3} \cite{Howard_2019_ICCV,Sandler_2018_CVPR} further widened the gap in accuracy between BNNs and compact networks.
Since this has reduced the practical applicability of BNNs, our goal in this work is to show that it is possible to achieve the milestone of \mobilenet{1}-level accuracy.

Prior work has been using full-precision architectures, \eg, AlexNet \cite{krizhevsky2012imagenet} and ResNet \cite{He2017}, without specific adaptations for BNNs.
To the best of our knowledge, only two works are exceptions:
Liu \etal added additional residual shortcuts to the ResNet architecture \cite{Liu_2018_ECCV} (the resulting model was reused in more recent work \cite{Gu_2019_ICCV,Martinez2020Training,Zhuang_2019_CVPR}) and Bethge \etal adapted a DenseNet architecture with dense shortcuts for BNNs \cite{Bethge_2019_ICCV_Workshops}.
In our work, we use another architectural approach \emph{MeliusNet} with designated building blocks to increase \emph{quality} and \emph{capacity} of features throughout the whole network (see \autoref{sec:melius-net}).
Further, a large share of operations in previous BNNs stems from a few layers which use 32-bit instead of 1-bit.
To solve this issue, we propose a redesign of these layers which saves operations and improves the accuracy at the same time (see \autoref{sec:deepstem-grouped}).

We evaluate MeliusNet on the ImageNet \cite{imagenet_cvpr09} dataset and compare it with the state-of-the-art (see \autoref{sec:results}).
To confirm the effectiveness of our methods, we also provided extensive ablation experiments.
During this study, we found that our training process with Adam \cite{kingma2014adam} achieves much better results than reported in previous work.
To allow for a fair comparison, we also trained the original (unchanged) networks and clearly separated the accuracy gains between the different factors in our ablation study (also within \autoref{sec:results}).
Finally, we conclude our work in \autoref{sec:conclusion}.

Summarized our main contributions in this work are:
\begin{itemize}
\itemsep0cm 
    \item A novel BNN architecture which counters the lower quality and lower capacity of binary feature maps efficiently.
    \item A more accurate and efficient initial set of grouped convolution layers for all binary networks.
    \item The first BNN that matches the accuracy of \mobilenet{1} 0.5, 0.75, and 1.0.
\end{itemize}
 
\section{Related Work}
\label{sec:related-work}

\noindent
Alternatives to binarization, such as
compact network structures \cite{Howard_2019_ICCV,Howard2017,Iandola2016,Sandler_2018_CVPR,Zhang2017},
knowledge distillation \cite{crowley2018moonshine,polino2018model,tung2019similarity}, and
quantized approaches \cite{courbariaux2015binaryconnect,Jung_2019_CVPR,Rastegari2016,Wang_2019_CVPR,Zhou2016,zhang2018lq} have been introduced in the past.
In this section, we take a more detailed look at approaches that use BNNs with 1-bit weights and 1-bit activations.
These networks were originally introduced by Courbariaux \etal \cite{Courbariaux2016} with \emph{Binarized Neural Networks} and improved by Rastegari \etal who used channel-wise scaling factors to reduce the quantization error in their \emph{XNOR-Net} \cite{Rastegari2016}.
The following works tried to further improve the network accuracy, which was much lower than the accuracy of common 32-bit networks, with different techniques:

For instance, they modified the loss function (or added new loss functions) instead of using a simple cross-entropy loss to train more accurate BNNs \cite{Gu_2019_ICCV,Martinez2020Training,Wang_2019_CVPR}.

\emph{WRPN} \cite{Mishra2017} and Shen \etal \cite{Shen2019a} increased the number of channels for a better performance.
Their work only increases the number of channels in the convolutions and the feature maps, but does not change the architecture.

Another way to increase the accuracy of BNNs was presented by \emph{ABC\nobreakdash-Net} \cite{Lin2017} and \emph{GroupNet} \cite{Zhuang_2019_CVPR}.
Instead of using a single binary convolution, they use a set of $k$ binary convolutions to approximate a 32-bit convolution (this number $k$ is sometimes called the number of binary bases).
This achieves higher accuracy but increases the required memory and number of operations of each convolution by the factor $k$.
These approaches optimize the network \emph{within} each building block.

The two approaches most similar to our work are Bi\nobreakdash-RealNet \cite{Liu_2018_ECCV} and BinaryDenseNet \cite{Bethge_2019_ICCV_Workshops}.
They use only a single binary convolution, but adapt the network architecture compared to full-precision networks to improve the accuracy of a BNN.
However, they did not test whether their proposed architecture changes are specific for BNNs or whether they would improve a 32-bit network as well.
 
\section{MeliusNet}
\label{sec:melius-net}

\noindent
The motivation for MeliusNet comes from the two main disadvantages of using binary values instead of 32-bit values for weights and inputs.

On the one hand, the number of possible \emph{weight} values is reduced from up to $2^{32}$ to only $2$.
This leads to a certain quantization error, which is the difference between the result of a regular 32-bit convolution and a 1-bit convolution.
This error reduces the \emph{quality} of the features computed by binary convolutions compared to 32-bit convolutions.

On the other hand, the value range of the \emph{inputs} (for the following layer) is reduced by the same factor.
This leads to a huge reduction in the available \emph{capacity} of features as well, since fine-granular differences between values, as in 32-bit floating point values, can no longer exist.

In the following section, we describe how MeliusNet increases the quality and capacity of features efficiently.
Afterwards, we describe how the number of operations in the remaining 32-bit layers of a binary network can be reduced.
Finally, we show the implementation details of our BNN layers.

\begin{figure}[t]
\captionsetup[subfigure]{justification=centering}
\begin{subfigure}[t]{0.59\linewidth}
    \vskip 0pt
    \centering
    \includegraphics[width=0.9\linewidth]{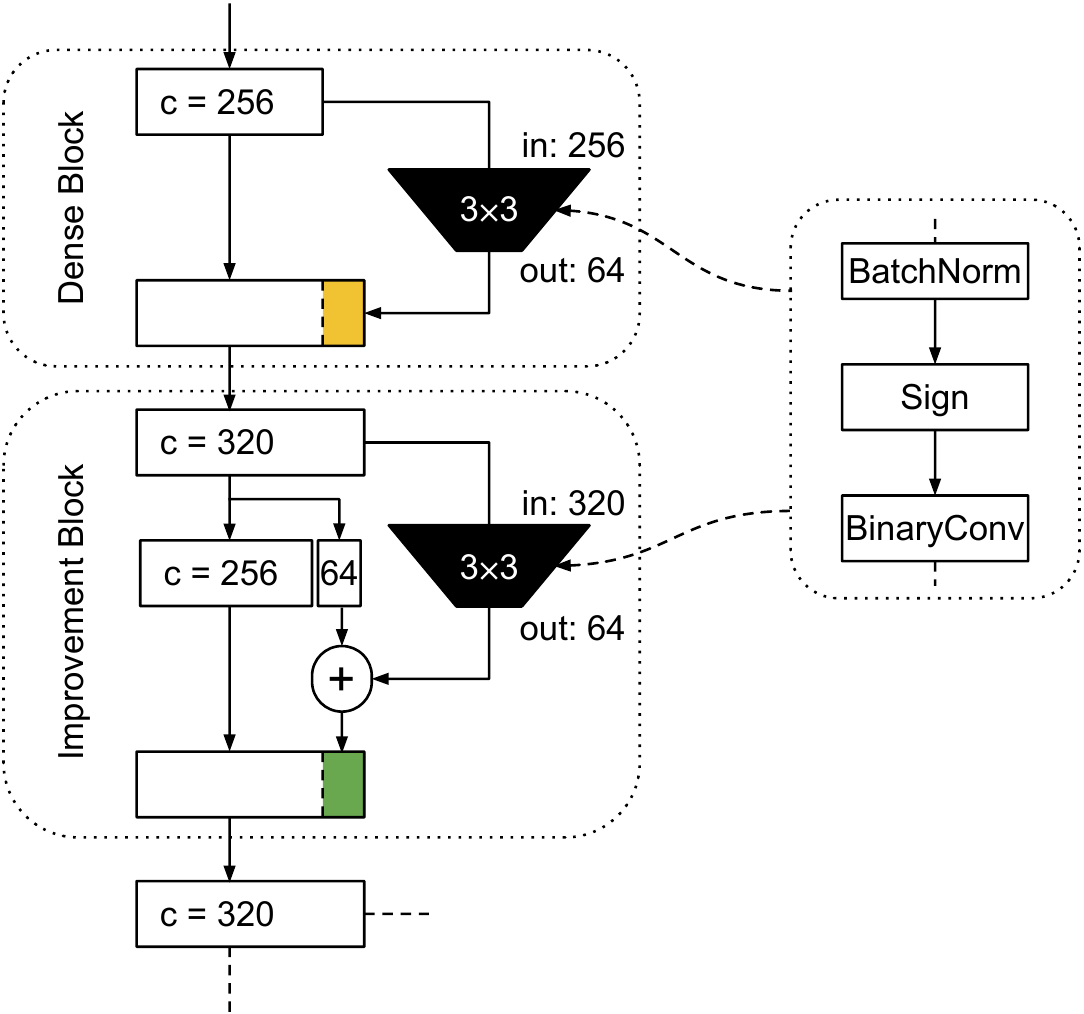}
    \caption{
      Building blocks of MeliusNet (c denotes the number of channels in the feature map).
      We first increase the feature \emph{capacity} by concatenating 64 newly computed channels to the feature map (yellow area) in the Dense Block.
      Then, we improve the \emph{quality} of \emph{those newly added channels} with a residual connection (green area) in the Improvement Block.
      The result is a balanced increase of \emph{capacity} and \emph{quality}.
    }
    \label{fig:meliusnet-structure}
\end{subfigure}
\begin{subfigure}[t]{0.40\linewidth}
    \vskip 0pt
    \centering
    \includegraphics[width=0.98\linewidth]{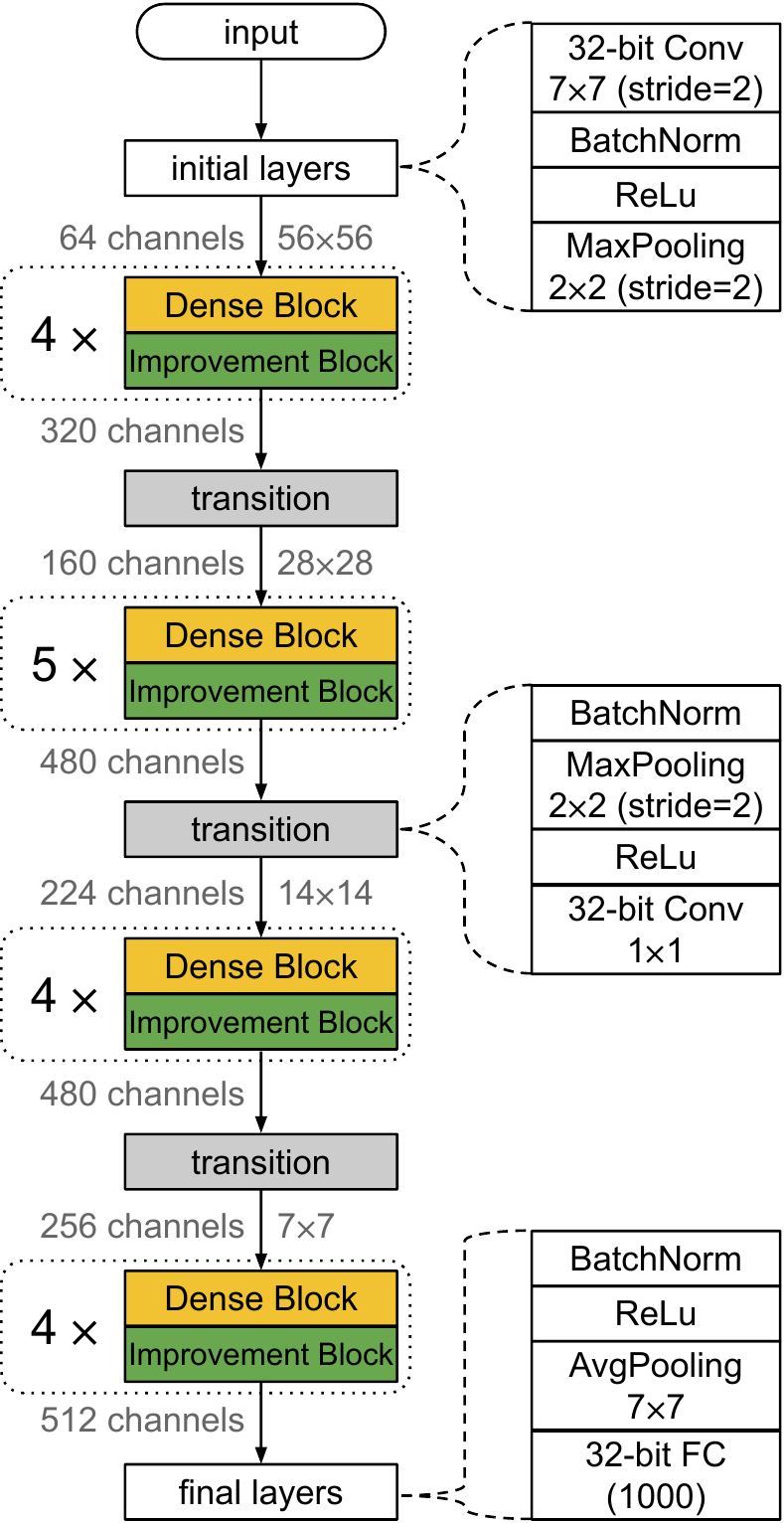}
    \caption{
    A depiction of our MeliusNet 22 with a configuration of 4-5-4-4 blocks between transitions.
    }
    \label{fig:melius-net-22}
\end{subfigure}
\caption{The building blocks and an exemplary network structure of MeliusNet.}
\label{fig:melius-net}
\end{figure}

\subsection{Increasing Capacity and Improving Quality}
\label{sec:main-melius-net}

\noindent
The core building block of MeliusNet consists of a \emph{Dense Block} followed by an \emph{Improvement Block} (see \autoref{fig:meliusnet-structure}).
The \emph{DenseBlock} increases feature \emph{capacity}, whereas the \emph{Improvement Block} increases feature \emph{quality}.

The \emph{Dense Block} was inspired by the BinaryDenseNet architecture \cite{Bethge_2019_ICCV_Workshops}, which is a binary variant of the DenseNet architecture \cite{Huang2016}.
It consists of a binary convolution which derives 64 channels of new features based on the input feature map, with, for example, 256 channels.
These features are concatenated to the feature map itself, resulting in 320 channels afterwards, thus increasing feature \emph{capacity}.

The novel \emph{Improvement Block} increases the quality of these newly concatenated channels.
It uses a binary convolution to compute 64 channels again based on the input feature map of 320 channels.
These 64 output channels are added to the previously computed 64 channels through a residual connection, without changing the first 256 channels of the feature map (see \autoref{fig:meliusnet-structure}).
Thus, this addition improves the last 64 channels, leading to the name of our network (\emph{melius} is latin for \emph{improvement}).
With this approach each section of the feature map is improved exactly once.

Note that we could also use a residual connection to improve the \emph{whole} feature map instead of using the proposed \emph{Improvement Block}.
However, with this naive approach, the number of times each section of the feature map is improved would be highly skewed towards the initially computed features.
It would further incur a much higher number of operations, since the number of output channels needs to match the number of channels in the feature map.
With the proposed \emph{Improvement Block}, we can instead save computations and get a feature map with balanced quality improvements.
Our experiments showed that using a regular residual connection instead of our \emph{Improvement Block} leads to $\sim$3\% \emph{lower} accuracy on ImageNet for equally sized networks (see the supplementary material for details).

As stated earlier, alternating between a \emph{Dense Block} and an \emph{Improvement Block} forms the core part of the network.
Depending on how often the combination of both blocks is repeated, we can create models of different size and with a different number of operations.
Our network progresses through four stages, with transition layers in between, which halve the height and width of the feature map with a MaxPool layer.
Furthermore, the number of channels is also roughly halved in the $1\times1$ downsampling convolution during the transition (see \autoref{tab:melius-details} on page \pageref{tab:melius-details} for the exact factors).
We show an example in \autoref{fig:melius-net-22}, where we repeat the blocks 4, 5, 4, and 4 times between transition layers and achieve a model which is similar to Bi\nobreakdash-RealNet18 \cite{Liu_2018_ECCV} in terms of model size.

\subsection{Creating an Efficient Stem Architecture}
\label{sec:deepstem-grouped}

\noindent
We follow previous work and do not binarize the first convolution, the final fully connected layer, and the $1\times1$ (``downsampling'') convolutions in the network to preserve accuracy \cite{Bethge_2019_ICCV_Workshops,Liu_2018_ECCV,Zhuang_2019_CVPR}.
However, since these layers contribute a large share of operations, we propose a redesign of the first layers.
We hypothesize that improving the first set of layers in an efficient way should generalize well to all BNN architectures.
Note that we refer to the ImageNet classification task \cite{imagenet_cvpr09} in the following examples.

We compared previous BNNs to the compact network architecture \mobilenet{1} 0.5 \cite{Howard2017}, which only needs $1.49\cdot10^8$ operations in total and can achieve $63.7\%$ accuracy on ImageNet.
We found, that the closest BNN result (regarding model size and operations) is Bi\nobreakdash-RealNet34, which achieves lower accuracy ($62.2\%$) with a similar model size, but it also needs more operations ($1.93\cdot10^8$).
We presume, that because of this difference, compact model architectures are more popular for practical applications than BNNs, especially with more recent (and improved) compact networks appearing \cite{Howard_2019_ICCV,Sandler_2018_CVPR}.
To find a way to close this gap, we analyze the required number of operations in the following.

The first $7\times7$ convolution layer in a Bi\nobreakdash-RealNet18 \emph{alone} needs $65\%$ ($1.18\cdot10^8$) of the total operations of the whole network. 
The three $1\times1$ downsampling convolutions account for another $10\%$ ($0.18\cdot10^8$) of operations.
Since in total about $75\%$ of all $1.81\cdot10^8$ operations are needed for these 32-bit convolutions, we focused on them to reduce the number of operations.

\begin{figure}[t]
\captionsetup[subfigure]{justification=centering}
\begin{center}
\begin{subfigure}[t]{0.49\linewidth}
   \centering
   \includegraphics[width=0.4\linewidth]{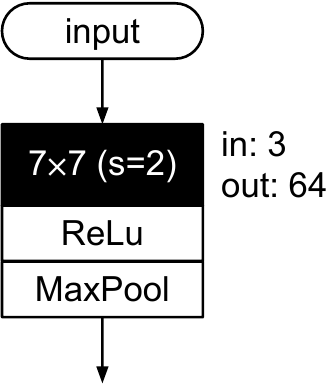}
   \caption{The $7\times7$ convolution with $1.18\cdot10^8$ operations.}
   \label{fig:7x7}
\end{subfigure}
\begin{subfigure}[t]{0.49\linewidth}
   \centering
   \includegraphics[width=0.4\linewidth]{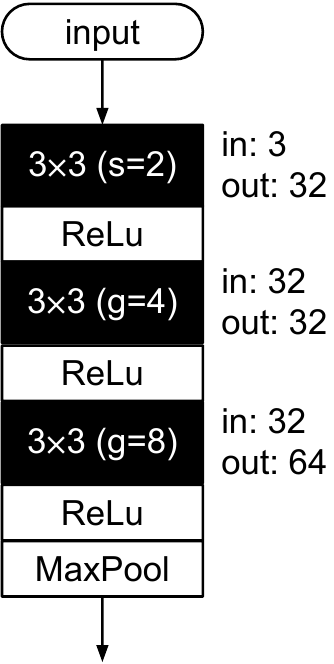}
   \caption{Our proposed \emph{grouped stem} with $0.69\cdot10^8$ operations.}
   \label{fig:grouped-stem}
\end{subfigure}
\end{center}
\caption{
A depiction of the two different versions of initial layers of a network (s is the stride, g the number of groups, we use 1 group and stride 1 otherwise).
Our \emph{grouped stem} in (\subref{fig:grouped-stem}) can be applied to all common BNN architectures, \eg, Bi-RealNet \cite{Liu_2018_ECCV} and BinaryDenseNet \cite{Bethge_2019_ICCV_Workshops}, as well as our proposed MeliusNet to save operations by replacing the expensive $7\times7$ convolution in the original layer configuration (\subref{fig:7x7}) without an increase in model size.
}
\label{fig:initial-layers}
\end{figure}
 
In previous work the $7\times7$ 32-bit convolution uses 64 channels.
We propose to replace the $7\times7$ convolution with three $3\times3$ convolutions, similar to the stem network used by Szegedy \etal \cite{szegedy2017inception}.
However, their stem network uses four times as many operations compared to the regular $7\times7$ convolution.
We use grouped convolutions \cite{krizhevsky2012imagenet} instead of regular convolutions for a reduction in operations (resulting in the name \emph{grouped stem}) and a lower number of channels.
The first convolution has 32 output channels (with a stride of 2), the second convolution uses 4 groups and 32 output channels, and the third convolution has 8 groups and 64 output channels (see \autoref{fig:initial-layers}).
This layer combination needs the same number of parameters (and thus model size) as the $7\times 7$ convolution, but only $0.69\cdot10^8$ instead of the original $1.18\cdot10^8$ operations, which is a reduction of more than $40\%$.

Similarly to adapting the stem structure, the $1\times1$ downsampling convolution can also use a certain number of groups, \eg, 2 or 4.
Since the features in the feature map are created consecutively with Dense Blocks, we add a channel shuffle operation before the downsampling convolution \cite{Zhang2017} (only if we use groups in our downsampling convolution).
This allows the downsampling convolution to combine features from earlier layers and later layers together.

Even though there are certainly other ways to change the 32-bit layers to reach an even lower number of operations, \eg, using quantization, a different set of layers, etc., our main goal is to highlight the high influence of these 32-bit layers on the number of operations in BNNs.
The 75\% share of operations in these layers was not clear to us in previous cost analyses of BNNs.
We hope this insight can direct future work into considering them for further improvement and investigate alternatives.
However, our proposed redesign already enables previous BNN works to reach a similar number of operations as \mobilenet{1} and we can test whether their accuracy can also achieve a similar level (see \autoref{sec:sota} for the results).

\subsection{Implementation Details}
\label{sec:general-binary}

\noindent
We follow the general principles to train binary networks as presented in previous work \cite{Bethge_2019_ICCV_Workshops,Liu_2018_ECCV,Rastegari2016}.
The weights and activations are binarized by using the \emph{sign} function:
\begin{equation}
    \mathrm{sign}(x) = \begin{cases} 
    +1 ~\text{if}~ x \geq 0, \\
    -1 ~\text{otherwise}.
    \end{cases}
\end{equation}
The non-differentiability of the sign function is solved with a Straight-Through Estimator (STE) \cite{BengioLC13_STE} coupled with gradient clipping as introduced by Hubara \etal \cite{Courbariaux2016}.
Therefore, the forward and backward passes can be described as:
\begin{align}
\label{eq:STE-forward}
    \text{Forward:}& ~r_o=\mathrm{sign}(r_i)~. \\
\label{eq:STE-backward}
    \text{Backward:}& ~\frac{\partial l}{\partial r_i}=\frac{\partial l}{\partial r_o}1_{|r_i|\leq t_\mathrm{clip}}~.
\end{align}
In this case $l$ is the loss, $r_i$ a real number input, and $r_o\in\{-1,+1\}$ a binary output.
We use a clipping threshold of $t_\mathrm{clip}=1.3$ as used by \cite{Bethge_2019_ICCV_Workshops}.
Furthermore, the computational cost of binary neural networks at runtime can be highly reduced by using the \texttt{xnor} and \texttt{popcount} CPU instructions, as presented by Rastegari \etal \cite{Rastegari2016}.

Previous work \cite{Liu_2018_ECCV} has suggested a different backward function to approximate the \emph{sign} function more closely, however, we found no performance gain during our experiments, similar to the results of \cite{bethge2018training}.
Channel-wise scaling factors have been proposed to reduce the difference between a regular and a binary convolution \cite{Rastegari2016}.
However, it was also argued, that they are mostly needed to scale the gradients \cite{Liu_2018_ECCV}, that a single scaling factor is sufficient \cite{Zhou2016}, or that neither of them is actually needed \cite{bethge2018training}.
Recent work suggests, that the effect of scaling factors might be neutralized by BatchNorm layers \cite{Bethge_2019_ICCV_Workshops}.
For this reason, and since we have not observed a performance gain by using scaling factors, we did not apply them in our convolutions.
We use the typical layer order (BatchNorm $\rightarrow$ sign $\rightarrow$ BinaryConv) of previous BNNs \cite{Bethge_2019_ICCV_Workshops,Liu_2018_ECCV}.
Finally, we replaced the bottleneck structures, consisting of a $1\times1$ and a $3\times3$ convolution, which are often used in full-precision networks, as it was done in previous work \cite{bethge2018training,Zhuang_2019_CVPR} and used a single $3\times3$ (1-bit) convolution instead.

\section{Results and Discussion}
\label{sec:results}

\noindent
We selected the challenging task of image classification on the ImageNet dataset \cite{imagenet_cvpr09} to test our new model architecture and perform ablation studies with our proposed changes.
Our implementation is based on BMXNet 2\footnote{\url{https://github.com/hpi-xnor/BMXNet-v2}} \cite{HPI_xnor} and the model implementations of Bethge \etal \cite{Bethge_2019_ICCV_Workshops}.
Note that experiment logs, accuracy curves, and plots of model structures for all trainings are in the supplementary material.

\begin{table}[t]
\caption{
Details of our different MeliusNet configurations, including the number of floating point and binary operations (FLOPs/BOPs), and their accuracy on the ImageNet classification task \cite{imagenet_cvpr09}.
The combined number of operations (OPs) is based on the speed-up factor of previous work \cite{Bethge_2019_ICCV_Workshops,Liu_2018_ECCV}: OPs = $\left(\frac{1}{64}\cdot\mathrm{BOPs}+\mathrm{FLOPs}\right)$.
The channel reduction factors are chosen at such specific fractions to keep the number of channels as multiples of 32.
The suffixes /2 and /4 denote, that the $1\times1$ downsampling convolutions use 2 and 4 groups, respectively.
}
\begin{small}
\begin{center}
\begin{tabular}{|l|c|c|r|c|c|c|c|}
\hline
\thead{Name\\(block numbers)} & \thead{Channel\\reduction factor\\in transitions} & \thead{BOPs\\$(\cdot10^9)$} & \thead{FLOPs\\$(\cdot10^8)$} &
\thead{OPs\\$(\cdot10^8)$} & \thead{Size\\(MB)} & \thead{OPs and Size\\similar to} & \thead{Top-1\\(Top-5)\\accuracy} \\ \hline
\mrc{2}{MeliusNet22\\(4,5,4,4)}      & \mrc{2}{$\frac{160}{320},\frac{224}{480},\frac{256}{480}$}  & \mrc{2}{4.62} & \mrc{2}{1.35} & \mrc{2}{2.08} & \mrc{2}{   3.9} & \mrc{2}{BDenseNet28\\\cite{Bethge_2019_ICCV_Workshops}} & \mrc{2}{63.6\%\\(84.7\%)} \\ & & & & & & & \\
\mrc{2}{MeliusNet29\\(4,6,8,6)}      & \mrc{2}{$\frac{128}{320},\frac{192}{512},\frac{256}{704}$}  & \mrc{2}{5.47} & \mrc{2}{1.29} & \mrc{2}{2.14} & \mrc{2}{   5.1} & \mrc{2}{BDenseNet37\\\cite{Bethge_2019_ICCV_Workshops}} & \mrc{2}{65.8\%\\(86.2\%)} \\ & & & & & & & \\
\mrc{2}{MeliusNet42\\(5,8,14,10)}    & \mrc{2}{$\frac{160}{384},\frac{256}{672},\frac{416}{1152}$} & \mrc{2}{9.69} & \mrc{2}{1.74} & \mrc{2}{3.25} & \mrc{2}{  10.1} & \mrc{2}{\mobilenet{1}\\0.75\cite{Howard2017}} & \mrc{2}{69.2\%\\(88.3\%)} \\ & & & & & & & \\
\mrc{2}{MeliusNet59\\(6,12,24,12)}   & \mrc{2}{$\frac{192}{448},\frac{320}{960},\frac{544}{1856}$} & \mrc{2}{18.3} & \mrc{2}{2.45} & \mrc{2}{5.25} & \mrc{2}{  17.4} & \mrc{2}{\mobilenet{1}\\1.0\cite{Howard2017}} & \mrc{2}{71.0\%\\(89.7\%)} \\ & & & & & & & \\ \hline
\mrc{2}{MeliusNetA\\(4,5,5,6)/4}     & \mrc{2}{$\frac{160}{320},\frac{256}{480},\frac{288}{576}$}  & \mrc{2}{4.85} & \mrc{2}{0.86} & \mrc{2}{1.62} & \mrc{2}{   4.0} & \mrc{2}{Bi-RealNet18\\\cite{Liu_2018_ECCV}} & \mrc{2}{63.4\%\\(84.2\%)} \\ & & & & & & & \\
\mrc{2}{MeliusNetB\\(4,6,8,6)/2}     & \mrc{2}{$\frac{160}{320},\frac{224}{544},\frac{320}{736}$}  & \mrc{2}{5.72} & \mrc{2}{1.06} & \mrc{2}{1.96} & \mrc{2}{   5.0} & \mrc{2}{Bi-RealNet34\\\cite{Liu_2018_ECCV}} & \mrc{2}{65.7\%\\(85.9\%)} \\ & & & & & & & \\
\mrc{2}{MeliusNetC\\(3,5,10,6)/4}    & \mrc{2}{$\frac{128}{256},\frac{192}{448},\frac{224}{704}$}  & \mrc{2}{4.35} & \mrc{2}{0.82} & \mrc{2}{1.50} & \mrc{2}{   4.5} & \mrc{2}{\mobilenet{1}\\0.5\cite{Howard2017}} & \mrc{2}{64.1\%\\(85.0\%)} \\ & & & & & & & \\
\hline
\end{tabular}\\
\end{center}
\end{small}
\label{tab:melius-details}
\end{table}

To compare to other state-of-the-art networks we created different configurations of MeliusNet with different model sizes and number of operations (see \autoref{tab:melius-details}).
Our main goal was to reach fair comparisons to previous architectures by using a similar model size and number of operations.
For example, we chose the configurations of MeliusNet22 and MeliusNet29 to be similar to BinaryDenseNet28 and BinaryDenseNet38, respectively.
Note that we calculated the number of operations in the same way as in previous work, factoring in a $64\times$ speed-up factor for binary convolutions \cite{Bethge_2019_ICCV_Workshops,Liu_2018_ECCV}.
For a comparison to Bi\nobreakdash-RealNet we further reduced the amount of operations, by using 2 or 4 groups in the downsampling convolutions for MeliusNetA and MeliusNetB, respectively and added a channel shuffle operation beforehand as described in \autoref{sec:deepstem-grouped}.
Finally, we created the networks MeliusNetC, MeliusNet42 and MeliusNet59 to be able to fairly compare to \mobilenet{1} 0.5, 0.75 and 1.0, respectively.
This also shows, that the basic network structure of MeliusNet can be adapted easily to create networks with different sizes and number of operations by tuning the number of blocks. 

\subsection{Grouped Stem Ablation Study and Training Details}
\label{sec:training-details-grouped-stem}

\begin{table*}[t]
\caption{
Ablation study on ImageNet \cite{imagenet_cvpr09} separating the gains between the training process and \emph{grouped stem}.
It shows the generic applicability of both.
}
\begin{small}
\begin{center}
\begin{tabular}{|c|c|c|c|c|c|c|c|}
\hline
\thead{Model\\size}  & \thead{Network\\Architecture} & \thead{Training\\procedure} & \thead{Grouped\\stem} &
\thead{OPs\\$(\cdot10^8)$} & \thead{Top-1\\acc.} & \thead{$\Delta$} & \thead{$\Delta$ of \\ method} \\ \hline
\mrc{8}{$\sim$4.0MB} & \mrc{3}{ResNetE18\cite{Bethge_2019_ICCV_Workshops}}        & Original & \xmark & $1.63$  & 58.1\%   & (base)   & $-1.9$    \\ 
                     &                                                            & Ours     & \xmark & $1.63$  & 60.0\%   & $+1.9$   & (base)    \\ 
                     &                                                            & Ours     & \cmark & $1.14$  & 60.6\%   & $+2.5$   & $+0.6$    \\ \cline{2-8}
                     & \mrc{3}{BinaryDenseNet28\cite{Bethge_2019_ICCV_Workshops}} & Original & \xmark & $2.58$  & 60.7\%   & $+2.6$   & $-1.0$    \\
                     &                                                            & Ours     & \xmark & $2.58$  & 61.7\%   & $+3.6$   & (base)    \\ 
                     &                                                            & Ours     & \cmark & $2.09$  & 62.6\%   & $+4.5$   & $+0.9$    \\ \cline{2-8}
                     & \mrc{2}{MeliusNet22 (ours)}                                & Ours     & \xmark & $2.57$  & 62.8\%   & $+4.7$   & (base)    \\ 
                     &                                                            & Ours     & \cmark & $2.08$  & 63.6\%   & $+5.5$   & $+1.1$    \\ \hline \hline
\mrc{8}{$\sim$5.1MB} & \mrc{3}{Bi-RealNet34\cite{Liu_2018_ECCV}}                  & Original & \xmark & $1.93$  & 62.2\%   & (base)   & $-1.1$    \\ 
                     &                                                            & Ours     & \xmark & $1.93$  & 63.3\%   & $+1.1$   & (base)    \\ 
                     &                                                            & Ours     & \cmark & $1.43$  & 63.7\%   & $+1.5$   & $+0.4$    \\ \cline{2-8}
                     & \mrc{3}{BinaryDenseNet37\cite{Bethge_2019_ICCV_Workshops}} & Original & \xmark & $2.71$  & 62.5\%   & $+0.3$   & $-0.8$    \\
                     &                                                            & Ours     & \xmark & $2.71$  & 63.3\%   & $+1.1$   & (base)    \\ 
                     &                                                            & Ours     & \cmark & $2.20$  & 64.2\%   & $+2.0$   & $+0.9$    \\ \cline{2-8}
                     & \mrc{2}{MeliusNet29 (ours)}                                & Ours     & \xmark & $2.63$  & 64.9\%   & $+2.7$   & (base)    \\ 
                     &                                                            & Ours     & \cmark & $2.14$  & 65.8\%   & $+3.6$   & $+0.9$    \\ \hline
\end{tabular}
\end{center}
\end{small}
\label{tab:ablation}
\end{table*}
 
\noindent
When training models with our proposed \emph{grouped stem} structure based on previous architectures, we discovered a large performance gain compared to previously reported results.
To verify the source of these gains we ran an ablation study on ResNetE18 \cite{bethge2018training} (which is similar to Bi\nobreakdash-RealNet18 \cite{Liu_2018_ECCV}, except for the addition of a single ReLu layer and a single BatchNorm), Bi\nobreakdash-RealNet34 \cite{Liu_2018_ECCV}, Binary\-Dense\-Net28/37\cite{Bethge_2019_ICCV_Workshops}, and our MeliusNet22/29 with and without our proposed grouped stem structure (see \autoref{tab:ablation}).

On the one hand, the results show, that using grouped stem instead of a regular $7\times7$ convolution increases the model accuracy for all tested model architectures.
The actual increase by using the grouped stem structure is between $0.4\%$ and $1.1\%$ for each model in addition to saving a constant amount ($0.49\cdot10^8$) of operations.
We conclude, that not only is using our grouped stem structure highly efficient, but it also generalizes well to different BNN architectures.

On the other hand, we also recognized that our training process performs significantly better than previous training strategies.
Therefore, we give a brief overview about our training configuration in the following:

For data preprocessing we use channel-wise mean subtraction, normalize the data based on the standard deviation, randomly flip the image with a probability of $\frac{1}{2}$ and finally select a random resized crop, which is the same augmentation scheme that was used in XNOR-Net \cite{Rastegari2016}.
We initialize the weights with the method of \cite{glorot2010understanding} and train our models from scratch (without pre-training a 32-bit model) for 120 epochs with a base learning rate of $0.002$.
We use the RAdam optimizer proposed by Liu \etal \cite{Liu2019b} and the default (``cosine'') learning rate scheduling of the GluonCV toolkit \cite{gluoncvnlp2019}.
This learning rate scheduling steadily decreases the learning rate based on the following formula ($t$ is the current step in training, with $0\leq t\leq 1$): $\mathrm{lr}(t)=\mathrm{lr}_\mathrm{base}\cdot\big(\frac{1 + \mathrm{cos}(\pi \cdot t)}{2}\big)$.
However, we achieved similar (only slightly worse) results with the same learning rate scheduling and the Adam \cite{kingma2014adam} optimizer, if we use a warm-up phase of 5 epochs in which the learning rate is linearly increased to the base learning rate.
Using SGD led to the worst results overall and even though we did some initial investigation into the differences between optimizers (included in supplementary material) we could not find a clear reason for the performance difference (a similar observation was made by Alizadeh \etal \cite{alizadeh2018empirical}).

\subsection{Ablation Study on 32-bit Networks}
\label{sec:ablation-32-bit}

\noindent
We performed another ablation study to find out whether our proposed MeliusNet is indeed specifically better for a BNN or whether it would also increase the performance of a 32-bit network.
Since our proposed MeliusNet without the Improvement Blocks is similar to a DenseNet, we compared these two architectures and trained two 32-bit models based on a DenseNet and a MeliusNet.
We used the off-the-shelf Gluon-CV training script for ImageNet and their DenseNet implementation as a basis for our experiment \cite{gluoncvnlp2019}.
To achieve a fair comparison, we constructed two models of similar size and operations.
We used 4-4-4-3 blocks (Dense Block and Improvement Block) between the transition stages for MeliusNet and 6-6-6-5 blocks (Dense Blocks only) for a DenseNet.
The models need 4.5 billion FLOPs with 20.87 MB model size and 4.0 billion FLOPs with 19.58 MB model size, respectively.
Therefore, we expect MeliusNet to definitely achieve a slightly better result, since it uses slightly more FLOPs and has a higher model size, unless our designed architecture is only specifically useful for BNNs.
Both models were trained with SGD with momentum ($\mathrm{lr} = 0.1$) and equal hyperparameters for 90 epochs (with a warm-up phase of 5 epochs and ``cosine'' learning rate scheduling).
Note that additional augmentation techniques (HSV jitter and PCA-based lightning noise) were used (in this study only), since we did not change the original Gluon-CV training script for the 32-bit models.

\begin{figure}[t]
\begin{center}
   \includegraphics[width=0.6\linewidth]{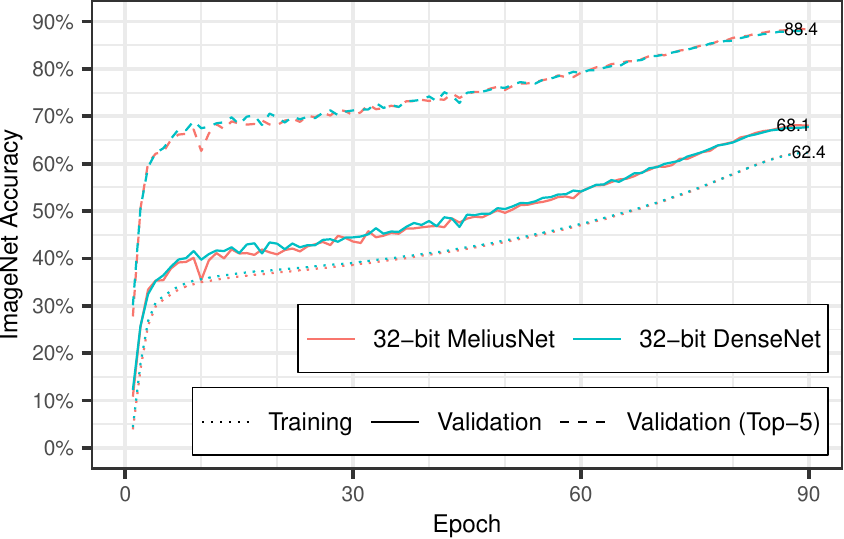}
\end{center}
   \caption{
   A comparison between the 32-bit versions of MeliusNet and DenseNet (best viewed in color).
   We tuned the number of building blocks to achieve models of similar complexity: MeliusNet uses 4,4,4,3 (4.5$\times10^{9}$ FLOPs, 20.87 MB) and DenseNet uses 6,6,6,5 (4.0$\times10^{9}$ FLOPs, 19.58 MB).
   We used the off-the-shelf Gluon-CV training script for ImageNet \cite{gluoncvnlp2019} with identical hyperparameters to train both models.
   The accuracy curves are almost indistinguishable for the whole training process and our 32-bit MeliusNet is not able to improve the result compared to a 32-bit DenseNet, even though it uses slightly more FLOPs and memory.
   }
\label{fig:fp-melius}
\end{figure}

The result shows basically identical training curves between both models for the whole training (see \autoref{fig:fp-melius}).
At the end of training, the training accuracy is even between both architectures at $62.4\%$.
Even though the validation accuracy does not match for the whole training, this is probably caused by randomized augmentation and shuffling of the dataset.
The accuracy gain of 1\% to 1.6\% that could be observed between a binary DenseNet and a binary MeliusNet (see \autoref{tab:ablation}) does not occur for the 32-bit version of networks.
Therefore, we conclude, that using our MeliusNet architecture for 32-bit models does not lead to an improvement, and our architecture is indeed only an improvement for BNNs.

\subsection{Comparison to State-of-the-art}
\label{sec:sota}

\begin{table*}[t]
\caption{
Comparison to state-of-the-art on ImageNet \cite{imagenet_cvpr09}.
All models were trained with cross-entropy loss.
Methods include low-bit quantization (first section) and approaches with multiple binary bases (second section).
The comparison is in parallel for two size categories (with differing number of layers).
The best result in each section is bold.
}
\begin{small}
\begin{center}
\begin{tabular}{|c|c|c|c|c|c|c|c|c|c|c|c|}
\hline
\mrc{3}{Method} & \mrc{3}{Bitwidth\\(W/A)} & \mcth{7}{ImageNet\vspace{-0.1cm}} \\ \cline{3-9}
& & \thead{Top-1\\Acc.} & \thead{Model\\Size} & \thead{OPs\\$(\cdot10^8)$} & & \thead{Top-1\\Acc.} & \thead{Model\\Size} & \thead{OPs\\$(\cdot10^8)$} \\ \hline
BWN\cite{Rastegari2016}                         & 1/32           & 60.8          &          4MB         & 18.1        && -             &           -           &  -         \\
TTQ\cite{zhu2016trained}                        & 2/32           & \textbf{66.6} &         5.3MB        & 18.1        && -             &           -           &  -         \\
HWGQ\cite{cai2017deep}                          & 1/2            & 59.6          &          4MB         & $\sim$2.4   && 64.3          &          5.1MB        & $\sim$3.4  \\
LQ-Net\cite{zhang2018lq}                        & 1/2            & 62.6          &          4MB         & $\sim$2.4   && \textbf{66.6} &          5.1MB        & $\sim$3.4  \\
SYQ\cite{faraone2018syq}                        & 1/2            & 55.4          &          4MB         & $\sim$2.4   && -             &           -           &  -         \\
DoReFa\cite{Zhou2016}                           & 2/2            & 62.6          &         5.3MB        & $\sim$2.4   && -             &           -           &  -         \\ \hline \hline
Ensemble\cite{Zhu_2019_CVPR}                    & (1/1)$\times$6 & 61.0          &           -          &  -          && -             &           -           &  -         \\
Circulant-CNN\cite{Cvpr2019}                    & (1/1)$\times$4 & 61.4          &           -          &  -          && -             &           -           &  -         \\
ABC-Net\cite{Lin2017}                           & (1/1)$\times$5 & 65.0          &        8.7MB         & 7.8         && -             &           -           &  -         \\
GroupNet\cite{Lin2017}                          & (1/1)$\times$5 & \textbf{67.0} &        9.2MB         & 2.68        && \textbf{70.5} &         15.3MB        & 4.13       \\ \hline \hline
BNN\cite{Courbariaux2016}                       & 1/1            & 42.2          & \mrc{09}{$\sim$4MB}  & 1.57        && -             & \mrc{09}{$\sim$5.1MB} &  -         \\
XNOR-Net\cite{Rastegari2016}                    & 1/1            & 51.2          &                      & 1.59        && -             &                       &  -         \\
Bi-RealNet\cite{Liu_2018_ECCV}                  & 1/1            & 56.4          &                      & 1.63        && 62.2          &                       & 1.93       \\
XNOR-Net++\cite{bulat2019xnor}                  & 1/1            & 57.1          &                      & 1.59        && -             &                       &  -         \\
Bi-RealNet\,(our baseline)                      & 1/1            & 60.6          &                      & 1.14        && 63.7          &                       & 1.43       \\
BDenseNet\cite{Bethge_2019_ICCV_Workshops}      & 1/1            & 60.7          &                      & 2.58        && 62.5          &                       & 2.71       \\
Strong Baseline\cite{Martinez2020Training}      & 1/1            & 60.9          &                      & 1.82        && -             &                       &  -         \\
BDenseNet\,(our baseline)                       & 1/1            & 62.6          &                      & 2.09        && 64.2          &                       & 2.20       \\
MeliusNetA,B\,(ours)                            & 1/1            & \textbf{63.4} &                      & 1.62        && \textbf{65.7} &                       & 1.96       \\ \hline \hline
32-bit baseline                                 & 32/32          & 69.3          &            46.8MB    & 18.1        && 73.3          &            87.2MB     & 36.6       \\ \hline
\end{tabular}\\
\end{center}
\end{small}
\label{tab:imagenet-sota}
\end{table*}
 
\noindent
The results of MeliusNetA and MeliusNetB compared to the state-of-the-art can be seen in \autoref{tab:imagenet-sota}.
Previous work has often compared two different size categories, BiRealNet18 and BiRealNet34 \cite{Liu_2018_ECCV}, without taking into account the cost in operations and model size.
For a proper cross-domain comparison to quantized approaches and approaches with multiple binary bases we included these numbers.
In those cases where the authors did not reveal the exact numbers, we calculated them to the best of our knowledge.
Since we trained the binary network architectures Bi-RealNet \cite{Liu_2018_ECCV} and BinaryDenseNet (BDenseNet) \cite{Bethge_2019_ICCV_Workshops} for our \emph{grouped stem} ablation study with our training strategy, we also report our results in addition to the accuracy reported by the original authors.

\subsubsection*{Comparison to other binary networks (one base):}

Overall, MeliusNetA and B achieve the best accuracy compared to other approaches with 1-bit activations and 1-bit weights without additional cost.
However, we recognize that by applying \emph{grouped stem} to the Bi-RealNet architecture it can also achieve a much lower cost than our MeliusNet, which could be useful for certain applications despite its lower accuracy.

We limited the comparison to other works that use cross-entropy loss as a training objective.
We note that Martinez \etal \cite{Martinez2020Training} showed that their approach with multi-stage knowledge distillation training can further enhance the accuracy and achieve 64.4\% over their 60.9\% accuracy of the ``Strong Baseline''.
However, their approach is orthogonal to ours, since we focus on the architectural improvement and purposely use only cross-entropy loss for training.
Similarly we have not included other work in \autoref{tab:imagenet-sota}, which uses more sophisticated training techniques, such as CI-Net \cite{Wang_2019_CVPR} and BONN \cite{Gu_2019_ICCV}.
These works achieved 59.9\% and 59.3\% accuracy on ImageNet (for a 4 MB model) with their improved training strategy, respectively.

\subsubsection*{Comparison to quantized networks:}

MeliusNet compares favorably to most quantized approaches (first section in \autoref{tab:imagenet-sota}), achieving a higher accuracy and lower resource cost than DoReFa \cite{Zhou2016} and HWGQ \cite{cai2017deep}.
Some quantized approaches can achieve a higher accuracy than MeliusNet, but they also require a significantly higher model size or higher number of operations.
TTQ \cite{zhu2016trained} with 2-bit weights and 32-bit activations achieves 66.6\% accuracy, but does not save any operations and has a higher model size.
LQ-Net \cite{zhang2018lq} achieves 0.9\% higher accuracy, but also needs about 75\% more operations.

\subsubsection*{Comparison to other binary networks (multiple bases):}

Comparing MeliusNetA and B to approaches with multiple bases (second section in \autoref{tab:imagenet-sota}) reveals that both ABC-Net \cite{Lin2017} and GroupNet \cite{Zhuang_2019_CVPR} achieve better results.
However, they come at a significant increase in model size and operations and represent a different approach of using multiple binary convolutions instead of a single binary convolution in each layer.
Still, the exceptionally high accuracy of GroupNet partly achieves the level of \mobilenet{1}, hence we examined it further in the next section with a comparison to the larger MeliusNet models.

\subsubsection*{Cross-domain comparison between BNNs and compact networks:}

For another challenging comparison, we compared our results based on Bi\nobreakdash-RealNet34, MeliusNetC, MeliusNet42, and MeliusNet59 to the compact network architecture \mobilenet{1} \cite{Howard2017} in \autoref{tab:comparison-mobile-net}.
Furthermore, we included the GroupNet approach \cite{Zhuang_2019_CVPR} as an alternative approach that uses 5 binary bases.

First of all, the comparison between \mobilenet{1} and MeliusNet shows small accuracy improvements between $0.4\%$ and $0.8\%$ across three different model sizes.
For a model size of $\leq$5.1 MB, a Bi-RealNet34 trained with \emph{grouped stem} also shows the potential to reach the same accuracy with a lower amount of operations.
This shows that our proposed \emph{grouped stem} structure can effectively reduce the gap between \mobilenet{1} and previous BNN work.

We note that the GroupNet approach can also achieve an accuracy similar to \mobilenet{1} 1.0, although they have not shown the same level of accuracy for smaller model sizes, \eg, \mobilenet{1} 0.5 and 0.75.
In addition, GroupNet and MeliusNet differ in their approach.
GroupNet replaced a single binary convolution with multiple ones while reusing a regular Bi-RealNet architecture, whereas MeliusNet uses a novel architecture but with a single binary convolution per layer.
This also means both approaches could be combined in future work to achieve even more accurate BNNs.

\begin{table}[t]
\caption{
Comparison of \mobilenet{1} \cite{Howard2017}, the GroupNet approach \cite{Zhuang_2019_CVPR}, which uses multiple binary bases, and our results, based on Bi-RealNet34 \cite{Liu_2018_ECCV} and our binary MeliusNet on the ImageNet dataset \cite{imagenet_cvpr09}.
We can achieve a similar or better accuracy than MobileNet 0.5, 0.75 and 1.0 with different networks.
}
\begin{small}
\begin{center}
\begin{tabular}{|c|c|c|c|c|c|c|c|}
\hline
\thead{Model\\size} & \thead{Architecture} & \thead{Bitwidth\\(W/A)} & \thead{OPs\\$(\cdot10^8)$} & \thead{Top-1\\acc.} & \thead{$\Delta$} \\ \hline
9.2MB          & GroupNet18 \cite{Zhuang_2019_CVPR}            & (1/1)$\times$5   & $2.68$  & 67.0\%          & -        \\
15.3MB         & GroupNet34 \cite{Zhuang_2019_CVPR}            & (1/1)$\times$5   & $4.13$  & 70.5\%          & -        \\ \hline
5.1MB          & Bi-RealNet34 \cite{Liu_2018_ECCV}             & 1/1              & $1.93$  & 62.2\%          & $-1.5$   \\ 
5.1MB          & \emph{\mobilenet{1} 0.5} \cite{Howard2017}    & 32/32            & $1.49$  & 63.7\%          & (base)   \\ 
5.1MB          & Bi-RealNet34$^*$ \cite{Liu_2018_ECCV}         & 1/1              & $1.43$  & 63.7\%          & $\pm0.0$ \\ 
4.5MB          & \textbf{MeliusNetC}                           & 1/1              & $1.50$  & \textbf{64.1\%} & $+0.4$   \\ \hline
\mrc{2}{10MB}  & \emph{\mobilenet{1} 0.75} \cite{Howard2017}   & 32/32            & $3.25$  & 68.4\%          & (base)   \\ 
               & \textbf{MeliusNet42}                          & 1/1              & $3.25$  & \textbf{69.2\%} & $+0.8$   \\ \hline
\mrc{2}{17MB}  & \emph{\mobilenet{1} 1.0} \cite{Howard2017}    & 32/32            & $5.69$  & 70.6\%          & (base)   \\ 
               & \textbf{MeliusNet59}                          & 1/1              & $5.32$  & \textbf{71.0\%} & $+0.4$   \\ \hline
\end{tabular}\\
{\small $^*$ This result is based on our training using grouped stem.}\\
\end{center}
\end{small}
\label{tab:comparison-mobile-net}
\end{table}
 
We conclude that MeliusNet is a valid alternative to the decomposition strategy described in GroupNet, since it is more flexible for creating models with different size and number of operations.
MeliusNet also shows very promising results to be comparable to \mobilenet{1} since it surpasses their accuracy for three different model sizes.

\section{Conclusion}
\label{sec:conclusion}

\noindent
Previous work has shown different techniques to increase the accuracy of BNNs by increasing the channel numbers or replacing the binary convolutions with convolutions with multiple binary bases.
The Bi\nobreakdash-RealNet and the BinaryDenseNet approaches were the first to change the architecture of a BNN compared to a 32-bit network.
In our work, we showed a novel architecture \emph{MeliusNet}, which is specifically designed to amend the disadvantages of using binary convolutions.
In this architecture, we repeatedly add new features and improve them to compensate for the lower quality and lower capacity of binary feature maps.
Our experiments with different model sizes on the challenging ImageNet dataset show that MeliusNet is superior to previous BNN approaches, which adapted the architecture.

Further, we presented \emph{grouped stem}, an optimized set of layers that can replace the first convolution.
This has considerably reduced the gap between BNN results and compact networks, and with our optimization, both previous architectures and our proposed MeliusNet can reach an accuracy similar to \mobilenet{1} 0.5, 0.75, and 1.0 based on the same model size and a similar amount of operations.
This provides a strong basis for BNNs to gain popularity and possibly achieve future milestones, such as reaching an accuracy similar to \mobilenet{2} or \mobilenetsuffix{3}.
The higher energy saving potential of BNNs (based on customized hardware) could then make them the favorable choice for many applications.

\clearpage
\bibliographystyle{splncs04}
\bibliography{library}

\clearpage
\appendix

\section*{Supplementary material}
\label{sec:contents}

\noindent
Our supplementary material contains the following information:

\begin{itemize}
    \item \autoref{sec:exp-details} briefly explains the structure of the experiment data, which can be found here: \url{https://owncloud.hpi.de/s/h5zWIepW1OS0Rs6}
    \item \autoref{sec:naive-vs-meliusnet} shows a comparison between MeliusNet and the naive approach of simply alternating Residual Blocks and Dense Blocks
    \item \autoref{sec:optimizers} contains data that shows some of the observed differences between the different optimizers (SGD, Adam, RAdam)
\end{itemize}
\vspace{0.6cm}

\section{Detailed Experiment Data}
\label{sec:exp-details}

\begin{figure*}[b]
\captionsetup[subfigure]{justification=centering}
\begin{center}
\begin{subfigure}[t]{0.49\linewidth}
   \centering
   \includegraphics[width=\linewidth]{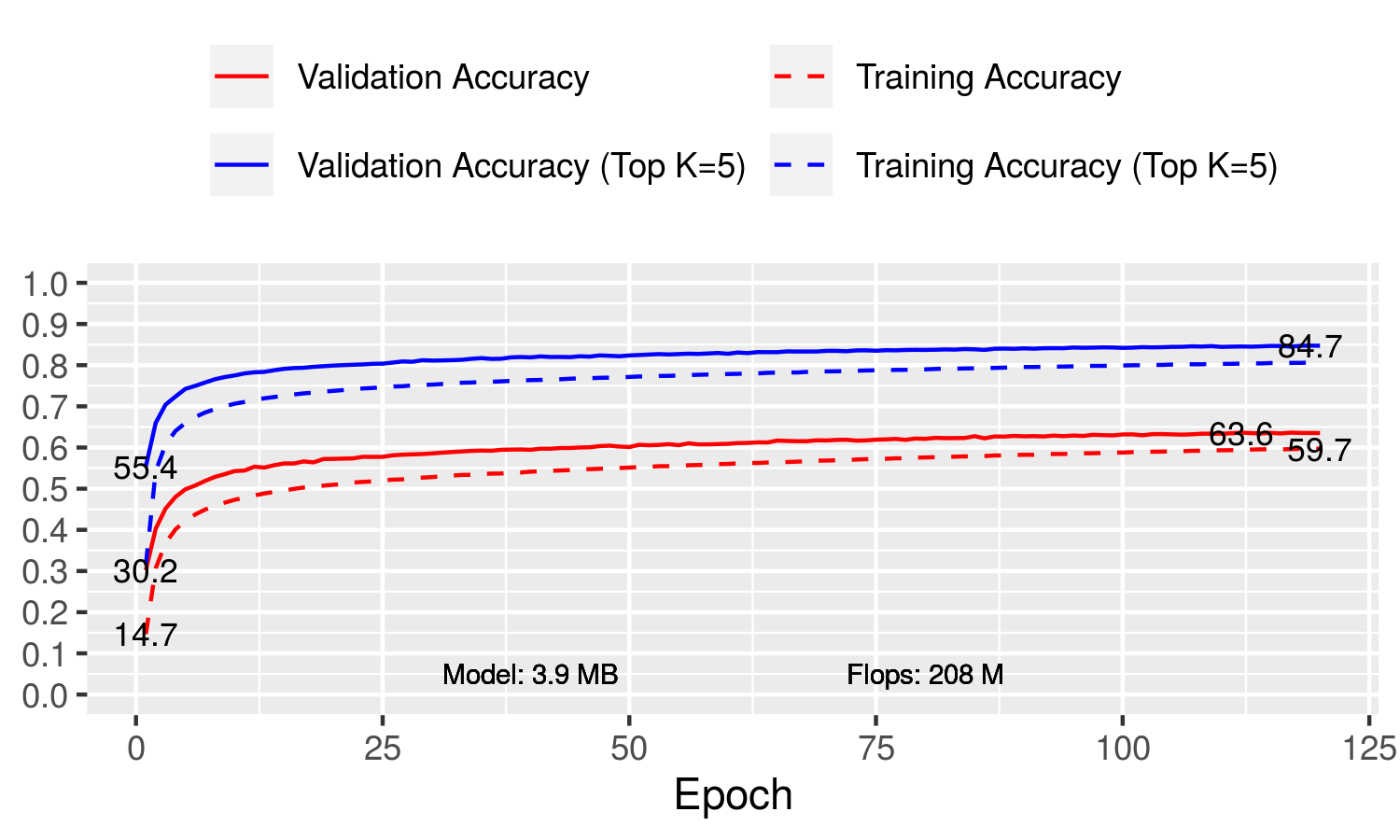}
   \caption{The accuracy curve.}
   \label{fig:example-accuracy}
\end{subfigure}
\begin{subfigure}[t]{0.49\linewidth}
   \centering
   \includegraphics[width=\linewidth]{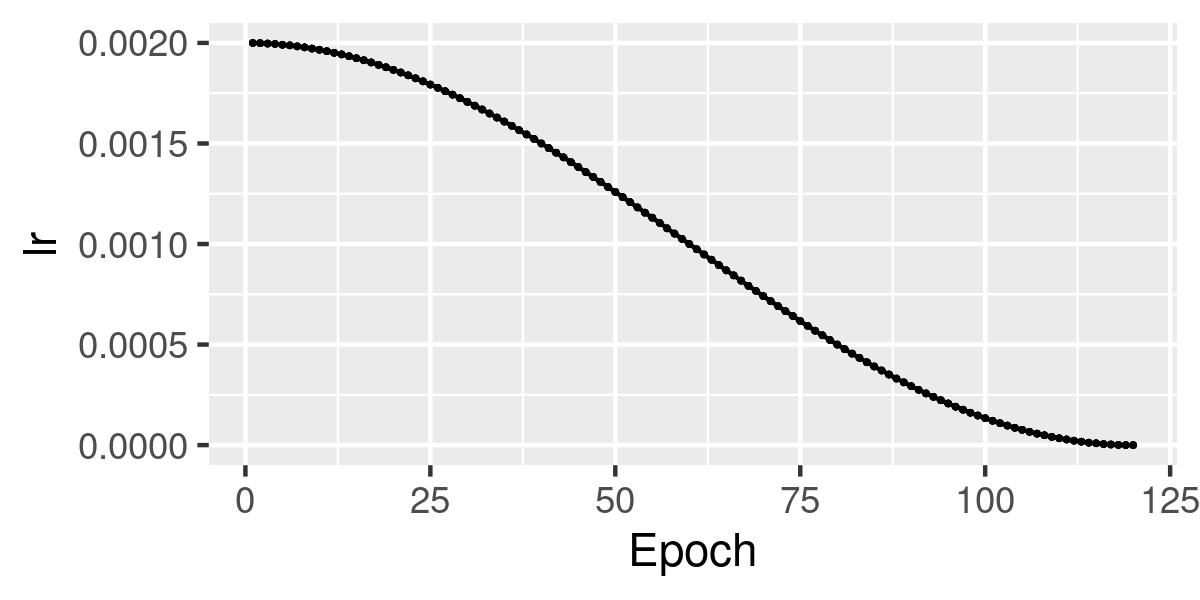}
   \caption{The ``cosine'' learning rate scheduling.}
   \label{fig:example-lr}
\end{subfigure}
\end{center}
\caption{
A visualization of the training process of MeliusNet22.
}
\label{fig:example-training}
\end{figure*}
 
\noindent
We include the experiment logs (\texttt{experiment.log}), accuracy curves (\texttt{accura\-cy.png}) and detailed plots (\texttt{network.pdf}) of our model architectures in one folder per experiment result within the parent folder ``main\_experiment\_data''.
The accuracy curves also include the model size and number of operations of the corresponding model.
An example of the accuracy curve of MeliusNet22 can be seen in \autoref{fig:example-training}.

\section{Comparing the Naive Approach and MeliusNet}
\label{sec:naive-vs-meliusnet}

\begin{figure*}[t]
\captionsetup[subfigure]{justification=centering}
\begin{center}
\begin{subfigure}[t]{0.49\linewidth}
   \centering
   \includegraphics[width=0.915\linewidth]{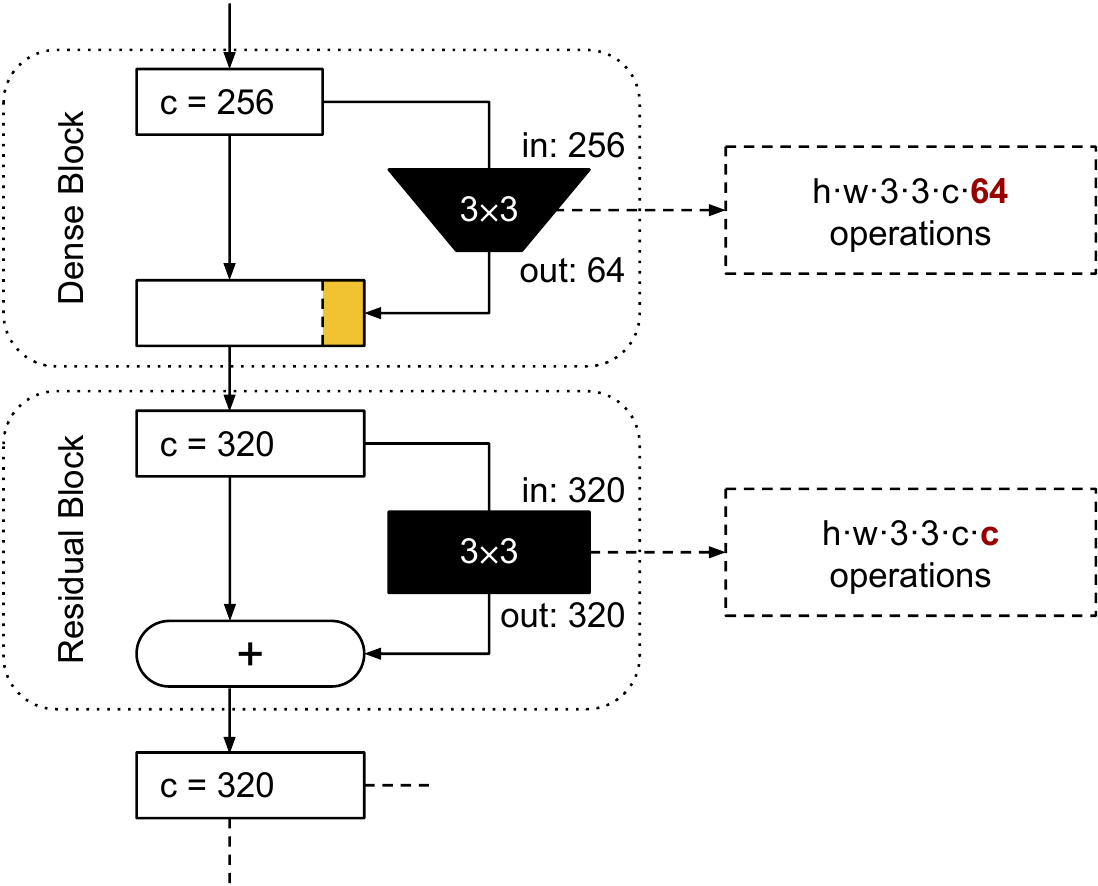}
   \caption{Naive approach}
   \label{fig:naive-structure-supp}
\end{subfigure}
\begin{subfigure}[t]{0.49\linewidth}
   \centering
   \includegraphics[width=0.915\linewidth]{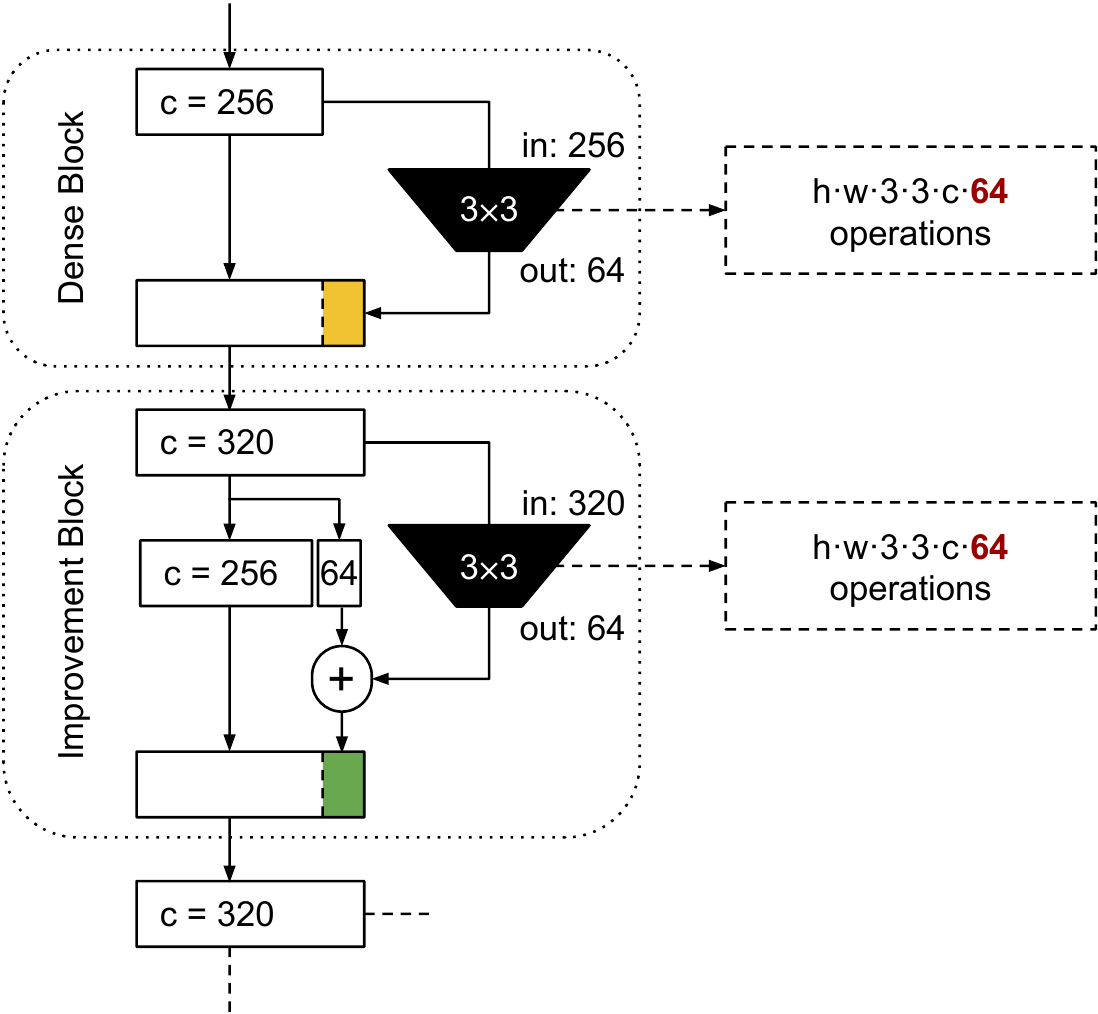}
   \caption{MeliusNet}
   \label{fig:meliusnet-structure-supp}
\end{subfigure}
\end{center}
\caption{
  The basic building blocks of MeliusNet and the naive approach of repeating alternating Dense Blocks and Residual Blocks ($c$ denotes the number of channels in the feature map).
  (\subref{fig:naive-structure-supp}) With the naive approach, the number of operations in the Residual Block increases by the factor of $c$ instead of the constant number of output channels (64) compared to the Dense Block.
  This means the Residual Block needs between 2 and 10 times the number of operations of the Dense Block, depending on the number of layers and depth of the layer in the network. 
  Furthermore the number of weights and operations increases quadratically, depending on $c$, making anything except very shallow networks unfeasible.
  (\subref{fig:meliusnet-structure-supp}) Our MeliusNet for comparison.
  The number of operations between both blocks is similar (only the number of input channels changes slightly between the blocks).
}
\label{fig:net-structures-supp}
\end{figure*}
 
\begin{figure}[t]
\begin{center}
   \includegraphics[width=0.7\linewidth]{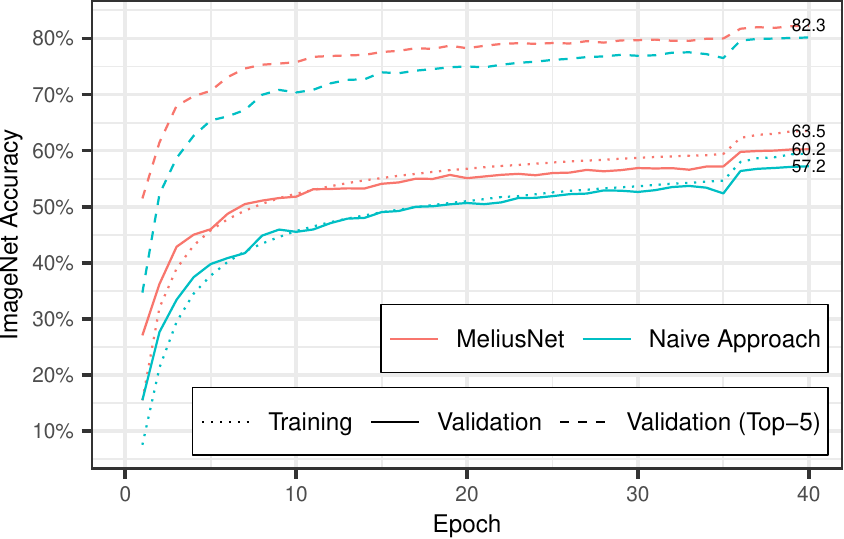}
\end{center}
   \caption{
   A comparison between MeliusNet and the naive approach of simply alternating a Residual Block and a Dense Block (see \autoref{fig:naive-structure-supp}).
   Both models need about 258 million operations (factoring in the speedup of binary operations).
   The 3\% accuracy drop of the naive approach is too large and the high number of operations needed for larger models make the naive architecture unfeasible for BNNs.
   }
\label{fig:naive-vs-melius}
\end{figure}
 
\noindent
The direct approach to combining residual and dense shortcut connections could lead to a result as shown in \autoref{fig:naive-structure-supp}.
In this case the combination of a Dense Block and a Residual Block are repeated throughout the network.
However, the residual shortcut connection requires that feature map sizes between the input and output of the convolution match.
This means the number of channel contributes to the number of of operations quadratically.
This makes achieving a reasonable number of operations difficult with this approach, since increasing the channel number (as is done in every Dense Block) leads to a quadratic increase of operations.
Therefore, increasing the capacity of feature maps with this approach is not practical, especially for larger binary networks.

\autoref{fig:meliusnet-structure-supp} shows the MeliusNet for comparison.
The design of our Improvement Block keeps the number of operations lower, since increasing the channel number with Dense Blocks only linearly increases the number of operations required for later blocks.

We also empirically evaluated both models.
These experiments were trained for only 40 epochs and a different learning rate schedule (base learning rate is $0.001$, decaying by $0.1$ at epochs 35 and 37).
However, since both models were trained with the same hyperparameters this should not affect the comparison between both.
Since we struggled to construct a model which could match in both model size and number of operations, we only made the number of operations equal.
In the comparison we can see that the naive approach is much worse, with a 3\% different in Top 1 accuracy on ImageNet (see \autoref{fig:naive-vs-melius}).
Even with the slightly smaller model (3.3 MB instead of 4 MB) this drop in accuracy is too much compared to other binary models, e.g. Bi-RealNet or BinaryDenseNet.
Therefore, we concluded that this approach is not useful for BNNs and have not pursued it further.
The details of these experiments are in the folder ``naive\_vs\_MeliusNet''.

\section{Optimizer Comparison}
\label{sec:optimizers}

\begin{figure*}[t]
\captionsetup[subfigure]{justification=centering}
\begin{center}
\begin{subfigure}[t]{0.49\linewidth}
   \centering
   \includegraphics[width=0.915\linewidth]{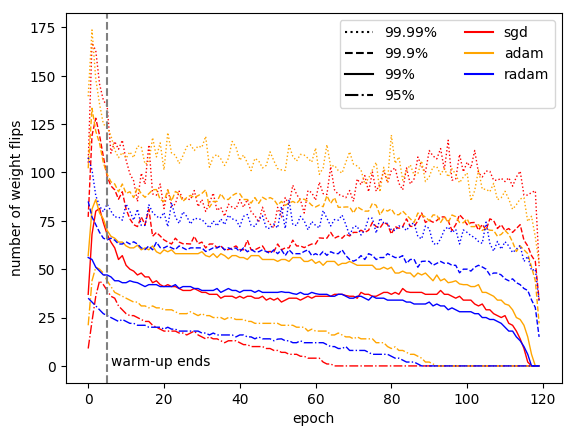}
   \caption{Data from the \emph{first} binary convolution of the \emph{first} network stage}
   \label{fig:percentiles-11}
\end{subfigure}
\begin{subfigure}[t]{0.49\linewidth}
   \centering
   \includegraphics[width=0.915\linewidth]{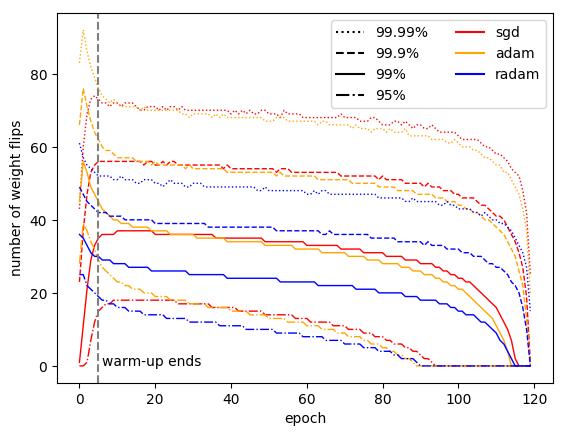}
   \caption{Data from the \emph{last} binary convolution of the \emph{last} network stage}
   \label{fig:percentiles-34}
\end{subfigure}
\end{center}
\caption{
We show the n-th percentile of the number of weight ``flips'' for each optimizer for the binary weights of two different convolution layers over the whole training process of 120 epochs for a ResNetE.
The first 5 epochs are warm-up epochs for Adam and SGD, where the learning rate is increased linearly to the base learning rate.
We can see, for example, that after the 100th epoch during a single given epoch 95\% of weights are stable in these layers.
Furthermore, for Adam and RAdam the stability increases during the training.
This is not the case for SGD in the earlier layers of the network (\eg, in (\subref{fig:percentiles-11})), where the number of flips increases starting around epoch 60.
}
\label{fig:percentiles}
\end{figure*}

\noindent
As written in the paper, we found, that both Adam and RAdam optimize better than SGD.
We tried different learning rates and learning rate schedules, however, the accuracy on ImageNet when training with SGD still was about 1\% lower than Adam (with warmup).
Therefore, we counted the number of sign ``flips'' for each individual weight between batches (accumulated per epoch) for each optimizer during the training of ResNetE18 on ImageNet (see \autoref{fig:percentiles}).
If a weight was updated from $-1$ to $+1$ when updating the weights after processing one batch its weight flip count would increase by one.
This can happen several times per epoch and intuitively reflects the ``stability'' of the training process regarding the binary weights.

First of all, the data showed, that surprisingly, after about 90 epochs, 95\% of all binary weights are stable during a single given epoch.
Note that this does not mean that 95\% of weights are stable for the \emph{whole time} after the 90th epoch, since the 95\% of stable weights are not necessarily identical between the different epochs.

During the training with Adam and RAdam, the average stability increases during the training, while for SGD the stability decreases after about 50 epochs.
However, this is only true for the earlier layers in the network (see \autoref{fig:percentiles-11}), but does not apply to later layers (see \autoref{fig:percentiles-34}).
Although this is an indication for a more unstable training process with SGD it does not yet conclusively explain the performance difference to RAdam and Adam.

\end{document}